\documentclass[letterpaper,10 pt,conference]{ieeeconf}
\IEEEoverridecommandlockouts
\usepackage{hyperref}
\usepackage{amsmath,amssymb,textcomp}
\usepackage{graphicx,booktabs,multirow,url,cite}
\usepackage{tikz}
\usetikzlibrary{arrows.meta,positioning,calc}
\newcommand{\method}{\textup{Nav3R}}
\title{\LARGE\bf Nav3R: Streaming Scene Reconstruction for Embodied Understanding and Navigation}
\author{Anonymous Authors}

\author{
    \textbf{Jiaming Zhao}$^{*}$\quad
    \textbf{Ting Huang}$^{*}$\quad
    \textbf{Qingxiang Liu}$^{*}$\quad
    \textbf{Zeyu Zhang}$^{*\dag}$\quad
    \textbf{Hao Tang}$^{\ddag}$ \vspace{0.1cm}\\
    School of Computer Science, Peking University\\
    \small $^*$Equal contribution. $^\dag$Project lead.
    $^\ddag$Corresponding author: bjdxtanghao@gmail.com
}

\makeatletter
\IEEEaftertitletext{%
\begin{minipage}{\linewidth}
\centering
\includegraphics[width=0.93\linewidth]{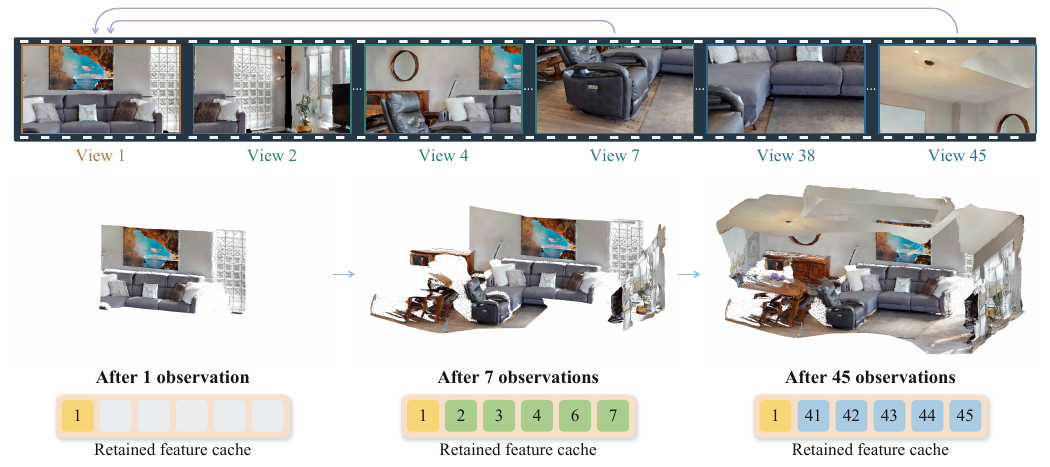}
\def\@captype{figure}%
\caption{\textbf{Nav3R} progressively reconstructs a persistent 3D scene from streaming observations, turning accumulated geometry into controlled memory for embodied understanding and navigation.}
\label{fig:sequence}
\end{minipage}\endgraf
}
\makeatother
\begin{document}
\maketitle
\thispagestyle{empty}
\pagestyle{empty}
\setcounter{dbltopnumber}{1} %

\begin{abstract}
Embodied understanding and navigation require persistent spatial memory to identify observed objects and guide actions across changing viewpoints. Learned streaming reconstruction can maintain such memory, but its geometry must be accurate enough for downstream use, and persistent reconstruction errors and irrelevant geometry can disrupt policy decisions. We present Nav3R, which converts streaming scene reconstruction into controlled memory for embodied understanding and navigation. A separate geometric-hint branch conditions causal streaming reconstruction on depth, validity, and confidence; offline hints come from sparse-view priors, while online deployment uses sensor depth directly. Geometrically consistent updates are admitted to a bounded memory that protects goal evidence, a fine-tuned point backbone proposes verified goal clusters, and a gate inserts a cluster into the frozen MTU3D policy only when it is persistent and more confident than the policy's own candidates. On held-out HM3D and 27 external sequences, the six-channel reconstruction front end reduces all six aggregate reconstruction errors. For Embodied Understanding, pre-registered confirmations on new episodes raise the category-level overlap between designated and observed goal surfaces by 2.66 points on fixed trajectories and by 3.69 points during active exploration. On all 1,000 HM3D ObjectNav-v2 validation episodes, Nav3R raises success at the standard 0.10\,m criterion from 65.4\% to 69.0\% ($+3.60$ points, 95\% CI $[+1.71,+5.77]$) and SPL by 1.47 points. Current-scene controls remove both gains. Direct accumulation of noise-free simulator depth, a privileged geometric reference, matches Nav3R in navigation (69.4\%) but designates more goal surface, so Nav3R attains about half of the reference understanding gain. These results support current-scene geometry as useful memory for embodied understanding and navigation under a frozen policy.
\href{https://github.com/AIGeeksGroup/Nav3R}{Code} \&
\href{https://aigeeksgroup.github.io/Nav3R}{Website}.
\end{abstract}

\section{INTRODUCTION}
\label{sec:introduction}
An embodied agent must identify language-specified objects and navigate toward them while its observations change continuously with viewpoint. A target may be visible through a doorway, leave the field of view after a turn, and become relevant again several steps later. Effective grounding and navigation therefore require memory that preserves useful spatial evidence across observations. Figure~\ref{fig:sequence} illustrates the resulting persistent scene representation. Existing ObjectNav systems use geometric, object-centric, or query-conditioned memory to support exploration and decision making~\cite{chaplot2020objectnav,ramakrishnan2022poni,yokoyama2024vlfm,zhu2025mtu3d}.

Causal reconstruction maintains a learned geometric state across views~\cite{liu2024slam3r,wang2025cut3r,lan2025stream3r}, so its errors can persist after surfaces leave the current view. Whether learned geometry is accurate enough for downstream use must therefore be tested against a reference. With known intrinsics and pose, noise-free simulator depth can be unprojected and accumulated into a metric point cloud without learned reconstruction; passing this accumulation through the same downstream branch provides such a reference, and the gap to it measures what residual reconstruction error costs each downstream task. A second challenge is reliability at the reconstruction-to-navigation interface. A misregistered update may persist in memory, irrelevant geometry may produce false object evidence, and uncertain point evidence may override a decision that the frozen policy would otherwise make correctly. Reconstruction quality alone is therefore insufficient; the memory and its policy interface must control which geometry is retained, which evidence forms an object hypothesis, and when that hypothesis can affect the decision.

These challenges lead to two design requirements. First, the system should build persistent current-scene geometry without retraining the downstream navigation policy. Second, reconstructed geometry should influence navigation only when its quality and semantic evidence support that use. Nav3R addresses these requirements with three stages: sparse-view guided streaming reconstruction, scene-aware 3D memory, and reconstruction-guided understanding and navigation. This separation also lets direct RGB-D accumulation serve as a geometric reference for learned reconstruction under the same downstream policy interface. The same proposal interface supports category-conditioned semantic grounding and navigation. The Embodied Understanding experiment tests whether the deployed system designates the observed surfaces of target objects, separately from its ability to reach a navigation goal.

Our contributions are threefold:
\begin{itemize}
\item We develop a sparse-view guided streaming reconstruction pipeline that combines overlap-aware view selection, frozen sparse-view priors, and a separate depth/validity/confidence hint path for causal propagation.
\item We introduce scene-aware 3D memory with transactional admission and goal-protected retention, together with point-semantic goal proposals that turn retained current-scene geometry into verified object hypotheses.
\item We connect the memory to a frozen ObjectNav policy through gated candidate insertion and evaluate reconstruction quality, Embodied Understanding, downstream navigation, memory properties, and interface safeguards under matched, pre-registered protocols.
\end{itemize}
\section{RELATED WORK}
\label{sec:related}
\noindent\textbf{Scene reconstruction.}
Classical SfM and SLAM recover motion and geometry through matching and optimization~\cite{schoenberger2016sfm,murartal2017orbslam2}. Recent learning-based methods directly predict geometric quantities, including pairwise point maps in DUSt3R and joint camera, depth, point-map, and track estimates in VGGT~\cite{wang2024dust3r,wang2025vggt}. NeuralRecon fuses monocular features volumetrically, SLAM3R couples local reconstruction with registration, and CUT3R, Spann3R, and STream3R maintain learned state across sequential views~\cite{sun2021neuralrecon,liu2024slam3r,wang2025cut3r,wang2025spann3r,lan2025stream3r}, while MVSplat provides sparse-view Gaussian priors~\cite{chen2024mvsplat}. Other recent work introduces camera/scene priors or alignment mechanisms~\cite{jang2025pow3r,chen2026ttt3r}. We reuse MVSplat for offline priors and STream3R for streaming reconstruction. These methods primarily target reconstruction accuracy, rendering quality, or geometric consistency; Nav3R instead evaluates learned current-scene geometry as persistent memory for embodied understanding and navigation.

\noindent\textbf{Embodied understanding and navigation.}
ObjectNav combines perception, spatial memory, exploration, and control to locate language-specified targets~\cite{chaplot2020objectnav,ramakrishnan2022poni,yokoyama2024vlfm}. Recent systems use semantic maps, visual-language features, object memories, and structured queries for long-horizon search~\cite{yokoyama2024vlfm,zhu2025mtu3d,wu2026softnav,liu2025navr1}. Habitat and HM3D provide standard evaluation environments~\cite{savva2019habitat,ramakrishnan2021hm3d}. We fine-tune Sonata-pretrained PTv3~\cite{wu2025sonata,wu2024ptv3} for point semantics. Nav3R studies this reconstruction-to-navigation interface through controlled memory admission, point-semantic goal proposals, and gated candidate insertion.
\section{METHOD}
\label{sec:method}
\subsection{Problem Formulation}
An ObjectNav episode consists of an environment, an initial agent state, and a language query $g$ specifying the target category. At time $t$, the agent observes RGB-D data, camera intrinsics, and sensor pose. The frozen MTU3D navigation policy maintains object and frontier memory and produces candidate-level decisions. We compare two auxiliary geometric memories: direct RGB-D accumulation and learned current-scene reconstruction. Let $\mathcal{M}_t$ denote the persistent auxiliary point memory and $\pi_0$ the frozen policy. Candidate generation, occupancy mapping, object memory, and low-level control remain unchanged. Geometric consistency before commitment, goal-aware retention, and gated policy influence govern $\mathcal{M}_t$. The downstream policy represents candidates by states $h_{t,i}$: its Decision head selects a candidate class, and its Ground head ranks candidates within that class. The policy and both heads remain frozen; Nav3R can only append one object candidate derived from $\mathcal{M}_t$, and otherwise the frozen outputs are used unchanged. For semantic grounding, the designated object is the inserted candidate when it is accepted and the highest-scoring frozen object candidate otherwise; the Embodied Understanding experiment compares its 3D support with the goal surfaces observed up to the current state. The same proposals support navigation through the object/frontier decision; no semantic-instance labels enter either policy input.

\subsection{Framework Overview}
\begin{figure*}[t]
\vspace*{0.6pt}%
\centering
\includegraphics[width=\textwidth]{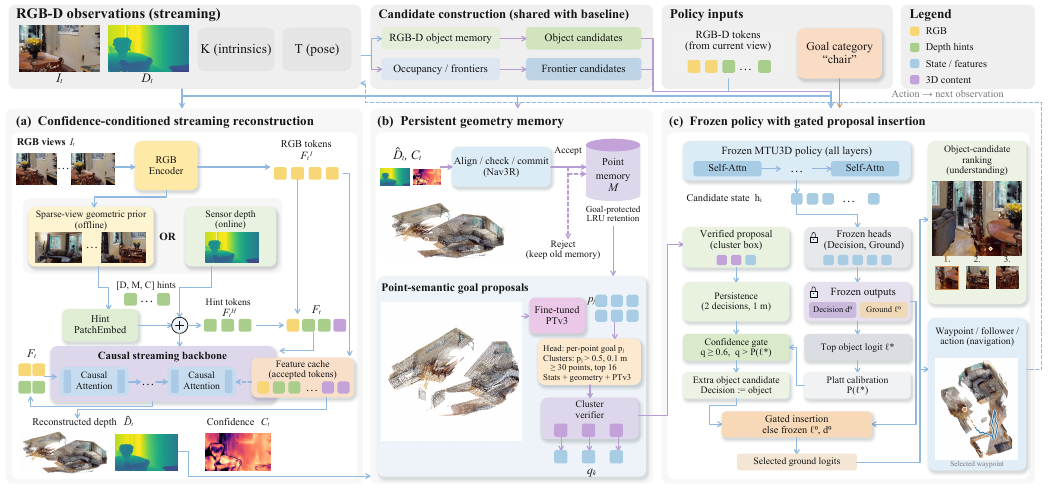}
\caption{\textbf{Overview of Nav3R.} Alternative offline and online hints feed streaming reconstruction. Checked, goal-protected point memory supplies point-semantic goal proposals; a verified proposal that passes the gate is inserted into the frozen policy's object candidates; actions produce new observations. The task outputs require separate evaluation.}
\label{fig:overview}
\vspace{-0.5cm}
\end{figure*}

As shown in Fig.~\ref{fig:overview}, Nav3R separates reconstruction from embodied understanding and navigation. Offline, overlapping sparse views provide a frozen Gaussian prior. A separate tokenizer converts projected depth, validity, and confidence into hints for the pretrained causal backbone, preserving the RGB embedding. Aligned points undergo transactional checks before entering persistent memory. Online ObjectNav uses sensor depth directly, bypassing view selection and sparse-view inference. Offline selection accesses the complete clip; only propagation is causal.

The frozen policy generates objects and frontiers from sensor observations. A fine-tuned point encoder assigns goal probabilities to memory points, connected clusters form object hypotheses, and a verifier scores them; a persistence and confidence gate decides whether the best cluster is inserted. Rejection leaves the frozen outputs unchanged at the same policy state.

\subsection{Two-Stage Scene Reconstruction}
\noindent\textbf{Sparse geometric hints.}
We select calibrated pairs that trade viewpoint variation against overlap: adjacent views are redundant, while distant views may not match reliably. For a calibrated clip $\mathcal S=\{I_t\}_{t=1}^{T}$, the candidate set contains every $1\leq i<j\leq T$. Let $b_{ij}=\lVert t_i-t_j\rVert_2$ denote the camera baseline and $o_{ij}$ the smaller cross-view project-valid ratio. Among the $K=16$ largest-baseline candidates, we choose
\begin{equation}
(i^*,j^*)=\operatorname*{arg\,max}_{\substack{(i,j)\in\mathcal C_K\\o_{ij}\geq0.18}} b_{ij}o_{ij},
\label{eq:view_selection}
\end{equation}
and use the largest-baseline pair if none passes the overlap threshold.

A frozen sparse-view predictor~\cite{chen2024mvsplat} estimates Gaussian means $\mu_k$, opacities $\alpha_k$, covariances $\Sigma_k$, and colors $c_k$ from the selected pair. We retain world-frame centers and weights satisfying
\begin{equation}
G_{ij}=\{(\mu_k,\alpha_k)\mid \alpha_k>0.2\},
\label{eq:sparse_prior}
\end{equation}
drop the farthest 1\% from their median, and deterministically subsample at most 120,000 points. Covariance and color are not passed to the streaming model. The nearest-visible projection instead converts the retained centers into sparse axial depth $D_t^h$, validity $M_t$, and clamped opacity $C_t\in[0,1]$. These three maps provide geometric location, support, and relative confidence for the next stage.

\noindent\textbf{Confidence-conditioned propagation.}
Nav3R uses a separate tokenizer for $X_t=[D_t^h,M_t,C_t]$, preserving the pretrained streaming backbone's RGB PatchEmbed~\cite{lan2025stream3r} while injecting geometric hints.

Let $F^{\rm rgb}_{t,p}$ be the frozen RGB token in patch $p$ and $H_{t,p}=\operatorname{LN}(E_h(X_t)_p)$ the token produced by the hint PatchEmbed. Confidence is pooled only over valid pixels, and the fused token is
\begin{align}
\bar c_{t,p}&=\frac{\sum_{j\in p}M_{t,j}C_{t,j}}{\sum_{j\in p}M_{t,j}+\epsilon}, \nonumber\\
F_{t,p}&=F^{\rm rgb}_{t,p}+\tanh(g)\,(2\bar c_{t,p})\,\mathbf1\!\left[\sum_{j\in p}M_{t,j}>0\right]H_{t,p}.
\label{eq:confidence_fusion}
\end{align}
Confidence controls both hint content and its patchwise contribution. Reconstruction training and benchmarking use projected sparse-view priors, whereas ObjectNav bypasses sparse-view inference at runtime and supplies valid Habitat depth with binary validity and confidence support. The hint branch is trained jointly with the camera and depth heads while the RGB aggregation, point-map head, and local correction remain frozen. The final reconstruction checkpoint is fixed before navigation evaluation. The learned scalar $g$ sets the global gate; empty patches receive no residual, and $C=0.5$ preserves its scale. Online, valid sensor depth gives $D_t^h=D_t$ with $M_t=C_t=1$; invalid pixels have zero support. Online confidence represents validity, not learned sensor uncertainty.

\subsection{Persistent Geometric Memory}
Navigation unprojects predicted axial depth using the known sensor intrinsics $K_t$ and pairs the resulting points with depth-head confidence. Let $\mathcal P_t^{\rm cv}=\{(p_j^{\rm cv},c_j)\}$, with $p_j^{\rm cv}=z_jK_t^{-1}[u_j,v_j,1]^\top$. Sensor depths in $[0.05,5.0]$\,m with positive prediction and confidence provide valid matches. We estimate the median depth ratio, then fit $D_j=s z_j+b$ on its central 80\%. The affine fit is retained only if it lowers the P90 residual, with $s\in(0.05,20)$ and $b\in(-2,2)$\,m. With $C=\operatorname{diag}(1,-1,-1)$ converting OpenCV to Habitat axes, the measured sensor pose $T_t=[R_t\mid t_t]$ gives
\begin{equation}
\widehat p_j^{\,w}=R_t C\!\left(\frac{s z_j+b}{z_j}p_j^{\rm cv}\right)+t_t.
\label{eq:camerachain}
\end{equation}
The filtered points define the world-frame proposal $\widehat{\mathcal P}_t$. Let $n_m$ count valid depth matches, $n_v$ trusted PTv3-input voxels, and $e_{50},e_{90}$ denote depth-residual quantiles. Geometric admission requires $n_m,n_v\geq128$, $e_{50}\leq0.10$\,m, and $e_{90}\leq0.20$\,m. Commitment additionally requires initialization or at least 0.05\,m subsequent translation:
\begin{equation}
\mathcal M_t=\begin{cases}
\operatorname{Merge}(\mathcal M_{t-1},\widehat{\mathcal P}_t),&a_t=1,\\
\mathcal M_{t-1},&a_t=0,
\end{cases}\quad a_t\in\{0,1\}.
\label{eq:transaction}
\end{equation}
Accepted points enter a 3\,cm voxel map capped at 131,072 points, retaining the highest-confidence representative, observation count, and last-seen step. When the cap binds, protected goal voxels are kept first and the remaining voxels are evicted from the least recently observed. The protected set holds at most 16,384 voxels whose goal probability (Sec.~\ref{sec:proposals}) exceeded 0.9 at a decision; a voxel leaves the set when its probability falls below 0.5, and the oldest entries leave first when the budget is exceeded. Rejected updates restore the prior map and cumulative step counter, preserve the accepted PTv3 cache, and reset the recurrent reconstruction session. Valid rotations may advance reconstruction without committing geometry; subsequent translation commits only its new proposal. A rejected observation may remain the older view of a later pair.

\subsection{Point-Semantic Goal Proposals}
\label{sec:proposals}
Accepted geometry within 12\,m is encoded in the cached agent frame after pruning below the fifth confidence percentile of the encoded points and 2\,cm highest-confidence voxelization, with at least 128 and at most 131,072 voxels. A Sonata-initialized PTv3~\cite{wu2025sonata,wu2024ptv3} fine-tuned for goal semantics produces 1,088-D features $f_j$, and a point head (LayerNorm and a $1088\!\rightarrow\!256\!\rightarrow\!256\!\rightarrow\!7$ GELU MLP over the six goal categories and background) gives the probability $p_j$ that point $j$ belongs to the query category $g$. Points with $p_j>0.5$ are grouped into 26-connected components on a 0.10\,m world-frame voxel grid; components with at least 30 points form proposals, of which the 16 with the highest mean probability are retained.

For proposal $\mathcal C_k$ with center $c_k$ and axis-aligned extent $e_k$, a 22-D descriptor $\phi_k$ contains the mean, maximum, and 10th/50th/90th percentiles of $p_j$, the log point count and log count of points with $p_j>0.9$, the three extents, the bottom, top, and center heights above the floor, the horizontal camera distance, overlap with any frozen object-candidate box, the distance to the nearest candidate center, and the category one-hot. With $\bar f_k$ the mean PTv3 feature of $\mathcal C_k$, the verifier scores
\begin{equation}
q_k=\sigma\big(\mathrm{MLP}\big([\phi_k;\,\mathrm{GELU}(W_p\,\mathrm{LN}(\bar f_k))]\big)\big),
\label{eq:verifier}
\end{equation}
where $W_p$ projects to 64-D and the MLP has two 256-unit GELU layers with dropout 0.2. The verifier predicts whether at least 30\% of the proposal's points lie on goal-category instances. Only $(c_k,e_k)$ and $q_k$ leave this stage; point features never enter the frozen policy.

\subsection{Navigation with Gated Insertion}
Decisions 0--3 follow the frozen baseline while the map matures. Let $k^*=\arg\max_k q_k$ be the best proposal at decision $t$, $\ell_t^\star$ the largest frozen Ground logit among valid object candidates, and $P(\ell)=\sigma(a\ell+b)$ its Platt calibration fitted on training decisions. A proposal is accepted only if it is confident, more confident than the frozen policy, and persistent over two consecutive decisions:
\begin{equation}
\begin{aligned}
A_t={}&\mathbf1[q_{k^*}\geq\theta_q]\,\mathbf1[q_{k^*}>P(\ell_t^\star)]\\[-1mm]
&\cdot\mathbf1\big[\lVert c_{k^*,t}-c_{k^*,t-1}\rVert_2\leq1\,\mathrm{m}\big],
\end{aligned}
\label{eq:gate_rule}
\end{equation}
where $c_{k^*,t-1}$ is the winning center at the previous decision and $\theta_q=0.6$. An accepted proposal is appended after the frozen object candidates, and the Decision output $d_t$ selects the object class; with $\ell^{\rm G}$ the Ground logits,
\begin{equation}
(\ell_t^{\rm G},d_t)=\begin{cases}
([\ell_t^{0,\rm G};\,\ell_{\max}],\,\mathrm{obj}),&A_t=1,\\
(\ell_t^{0,\rm G},\,d_t^{0}),&A_t=0,
\end{cases}
\label{eq:insertion}
\end{equation}
with $\ell_{\max}=100$, so that the inserted candidate wins its class and may be confirmed by the frozen geometric stopping rule. When the gate rejects the proposal, the exact frozen logits are used at the current policy state. A Habitat geodesic follower converts the selected waypoint into primitive actions without receiving the target coordinate. Rejection cannot undo earlier observations or trajectory differences; independent rollouts may produce different candidates.

\begin{table*}[!t]
\vspace*{5pt}%
\centering
\caption{\textbf{Scene-averaged Embodied Understanding results on pre-registered confirmation sets of new episodes.} Fixed trajectory: both systems queried at the same states of MTU3D's routes. Active: each system on its own route at its last natural decision. Surface metrics use 10\,cm matching on a 5\,cm grid; gain intervals are in the text.}
\vspace{-0.3cm}
\label{tab:understanding}
\footnotesize
\setlength{\tabcolsep}{3pt}
\resizebox{0.92\textwidth}{!}{%
\begin{tabular*}{\textwidth}{@{\extracolsep{\fill}}lcccccccc@{}}
\toprule
 & \multicolumn{4}{c}{Fixed trajectory (359 episodes, 1,334 states)} & \multicolumn{4}{c}{Active exploration (360 pairs, 313 with visible goal)} \\
\cmidrule(lr){2-5}\cmidrule(l){6-9}
Method & Category & Category & Category & Instance & Category & Category & Category & Instance \\
 & overlap (\%) $\uparrow$ & precision (\%) $\uparrow$ & recall (\%) $\uparrow$ & overlap (\%) $\uparrow$ & overlap (\%) $\uparrow$ & precision (\%) $\uparrow$ & recall (\%) $\uparrow$ & overlap (\%) $\uparrow$ \\
\midrule
MTU3D~\cite{zhu2025mtu3d} & 31.2 & 43.0 & 46.1 & 38.1 & 35.6 & 47.6 & 56.8 & 43.4 \\
\textbf{Nav3R} & \textbf{33.9} & \textbf{46.5} & \textbf{48.3} & \textbf{40.4} & \textbf{39.3} & \textbf{51.3} & \textbf{61.1} & \textbf{44.2} \\
\bottomrule
\end{tabular*}
}
\vspace{-0.5cm}
\end{table*}

\section{EXPERIMENTS}
\label{sec:experiments}
\subsection{Experimental Setup}
\noindent\textbf{Data.}
We use 10,267 synchronized $1920\!\times\!1080$ RGB observations with metric depth, intrinsics, and poses from 63 HM3D scenes in Habitat~\cite{savva2019habitat,ramakrishnan2021hm3d}: 51 for training, six for validation, and six for testing. Eighteen 7-Scenes~\cite{shotton2013seven} and nine Neural \mbox{RGB-D}~\cite{azinovic2022nrgbd} sequences test transfer.

Navigation and all ablations use the complete HM3D ObjectNav-v2 validation split: 1,000 paired episodes from 36 scenes and six target categories. The fixed-trajectory understanding experiment uses a development panel of 180 episodes (661 states with a visible goal) and a confirmation panel of 359 new episodes (1,334 states); the active experiment uses 396 development pairs and 360 new confirmation pairs. Confirmation sets reuse no earlier episode.

\begin{table}[t]
\caption{\textbf{Reconstruction errors for the binary-hint model and Nav3R.} The five-channel binary-hint model conditions the streaming backbone on RGB, depth hints, and a binary validity mask. Acc. and Comp. denote accuracy and completeness; lower is better. Metric errors are in meters, and raw errors use scale-normalized evaluator units. External results average 27 sequences.}
\vspace{-0.3cm}
\label{tab:reconstruction}
\centering
\footnotesize
\begin{tabular}{llrrr}
\toprule
Data & Metric & Binary & Nav3R & Gain \\
\midrule
\multirow{4}{*}{HM3D (6)} & Raw Acc. & 0.6485 & \textbf{0.6289} & 3.03\% \\
 & Raw Comp. & 1.0345 & \textbf{1.0031} & 3.03\% \\
 & Metric Acc. & 0.5029 & \textbf{0.4964} & 1.29\% \\
 & Metric Comp. & 1.0682 & \textbf{1.0422} & 2.44\% \\
\addlinespace[1.5pt]
\multirow{2}{*}{External (27)} & Raw Acc. & 0.4831 & \textbf{0.4658} & 3.58\% \\
 & Raw Comp. & 0.5215 & \textbf{0.4971} & 4.68\% \\
\bottomrule
\end{tabular}
\vspace{-0.6cm}
\end{table}

\noindent\textbf{Training.}
After a 100-step frozen-backbone stage adds the confidence channel, the 605,185-parameter hint branch and camera/depth heads follow a 500-step schedule with eight views, learning rate $5\!\times\!10^{-6}$, a 0.3 camera-head multiplier, and 10\% warm-up; we use the step-400 checkpoint, selected on validation. Relative-RMS regularization limits forgetting.

The point branch is trained on 600 HM3D training episodes from 145 scenes, captured with the deployed reconstruction under the evaluation protocol; semantic replay labels their 5,170 decision clouds, split by scene into training, selection, and confirmation partitions (3,703/725/742 clouds). The PTv3 backbone and the 348,551-parameter point head, initialized from a head trained on frozen Sonata features, train for two epochs with learning rates $2\!\times\!10^{-5}$ and $3\!\times\!10^{-4}$, raising selection goal mIoU from 0.462 to 0.546. The 160,193-parameter verifier trains on 3,841 proposals with three seeds; the seed with the best selection AUROC (0.936; 0.887 on confirmation) is kept. A pre-registered grid on the selection partition fixes $\theta_q$ and the persistence length by counterfactual rescued-minus-broken episodes, and the retention policy is selected on training episodes using an auxiliary proxy head outside Nav3R. Labels are computed after prediction and never enter policy inputs. Evaluations use frozen weights without retuning. All experiments are conducted on NVIDIA H100 GPUs with 80~GB of memory each.

\subsection{Scene Reconstruction}
\noindent\textbf{Protocol.}
The six-channel model adds continuous confidence and fine-tuning to its five-channel binary-hint parent; shared view selection and sparse-view priors isolate this combined change. Accuracy measures predicted-to-reference nearest-neighbor distance; completeness reverses that direction. Raw evaluation aligns median depth and median radial spread and measures GT-valid points in central $224\!\times\!224$ crops before ICP. Metric evaluation unprojects axial depth with supplied cameras and removes only the first-camera gauge, without ICP or Sim(3). We average sequences equally.

\noindent\textbf{Results.}
Table~\ref{tab:reconstruction} shows improvements in all six aggregate reconstruction errors. Gains range from 1.29\% to 3.03\% on held-out HM3D and from 3.58\% to 4.68\% on the 27 external sequences. Figure~\ref{fig:sequence} illustrates growing surface coverage over one camera sweep.
\subsection{Embodied Understanding}
\label{sec:understanding}

\begin{table*}[!t]
\vspace*{5pt}%
\caption{\textbf{ObjectNav results on all 1,000 paired episodes from the 36 HM3D validation scenes.} Success rates require STOP; Prox measures arrival without STOP. SPL and SoftSPL are fractions. All methods share the frozen policy, sensors, and action budget; the last row gives the 95\% scene-bootstrap interval of Nav3R minus MTU3D.}
\vspace{-0.3cm}
\label{tab:nav1000}
\centering
\footnotesize
\setlength{\tabcolsep}{3.5pt}
\resizebox{0.84\textwidth}{!}{%
\begin{tabular}{lccccccc}
\toprule
Method & SR@0.10 $\uparrow$ & SPL@0.10 $\uparrow$ & SR@0.25 $\uparrow$ & SPL@0.25 $\uparrow$ & Prox@0.25 $\uparrow$ & SoftSPL $\uparrow$ & DTG (m) $\downarrow$ \\
\midrule
MTU3D~\cite{zhu2025mtu3d} & 65.40\% & 0.1801 & 66.40\% & 0.1879 & 68.90\% & 0.2056 & 2.239 \\
Raw \mbox{RGB-D} memory & 69.40\% & 0.1976 & 70.50\% & 0.2037 & 71.90\% & 0.2187 & 2.089 \\
\addlinespace[1.5pt]
\textbf{\method{}} & 69.00\% & 0.1949 & 70.10\% & 0.2012 & 71.70\% & 0.2171 & 2.091 \\
\midrule
95\% CI ($\Delta$) & [+1.71,+5.77] & [0.0093,0.0211] & [+1.89,+5.86] & [0.0080,0.0197] & [+1.00,+4.98] & [0.0070,0.0165] & [\textminus{}0.267,\textminus{}0.049] \\
\bottomrule
\end{tabular}%
}
\vspace{-0.5cm}
\end{table*}

\noindent\textbf{Protocol and metrics.}
For Embodied Understanding, we score which object surfaces each system designates against official HM3D instance labels~\cite{yadav2023hm3dsem}. On a fixed trajectory, both systems are queried at the same states of MTU3D's routes, at the first decision of every 50-action window; MTU3D answers with the 3D support of its top-scoring object candidate and Nav3R with the cluster its deployment gate accepts at that state, otherwise with MTU3D's answer. Ground-truth semantic surfaces observed so far are rendered from the recorded poses and sampled on a 5\,cm grid; a reference point is selected if it lies within 10\,cm of a predicted point. Category overlap is the IoU between selected points and the observed surfaces of all goal-category instances; we also report its precision and recall, and an instance overlap against the best-matching goal instance. During active exploration, each system starts from the same state and is scored at its last natural decision against the surfaces that it observed; pairs in which both final states contain a visible goal enter the primary analysis.

Intervals use 10,000 paired scene-bootstrap draws and $p$-values a scene-level sign-flip test. Endpoints, samples, analysis scripts, and the decision rule ``CI lower bound $>0$'' were pre-registered; endpoints defined after a development look were confirmed once on new episodes.

\noindent\textbf{Results.}
On the fixed-trajectory confirmation panel, category overlap rises from 31.2\% to 33.9\% ($+2.66$ points, $[+1.88,+3.49]$, $p<10^{-5}$; Table~\ref{tab:understanding}), with precision and recall improving together ($+3.50$ and $+2.22$) and instance overlap by 2.34 points. The development panel gives $+2.39$ ($[+1.03,+4.13]$), and the effect holds at 5\,cm ($+1.34$) and 15\,cm ($+3.06$) matching. Chairs gain most (17.1\% to 23.2\%).
During active exploration, category overlap on the 360 new pairs rises from 35.6\% to 39.3\% ($+3.69$ points, $[+1.67,+5.79]$, $p=0.001$). The gain persists without the visibility condition ($+3.21$) and when both read-outs are compared at identical states of Nav3R's own route ($+3.71$), so it is not explained by easier trajectories alone.

The gains are category-level: under active exploration, instance overlap is not significantly improved ($+0.80$, $[-0.95,+2.59]$), since clusters can merge adjacent same-category instances. Earlier pre-registered read-outs showed no gain (fixed single-instance criterion $-3.93$; active strict single-instance success $-10.6$ and whole-cluster majority $+1.39$, $[-0.83,+3.89]$); development diagnosis traced this to off-surface reconstruction points and merged instances, motivating the surface-overlap metric before confirmation.

\begin{figure*}[!t]
\centering
\includegraphics[width=0.95\textwidth]{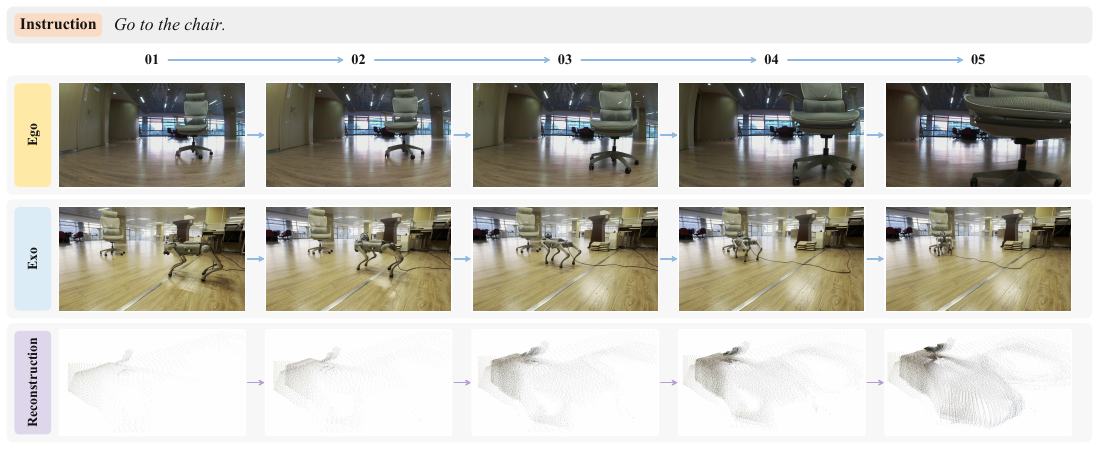}
\caption{\textbf{Qualitative real-world sequence of a quadruped robot approaching a chair.} Rows show egocentric RGB observations, external views, and point-cloud reconstruction snapshots. Columns show five chronological stages; arrows indicate temporal progression. The instruction summarizes the depicted task.}
\label{fig:realworld}
\vspace{-0.4cm}
\end{figure*}
\begin{figure}[!t]
\centering
\includegraphics[width=\columnwidth]{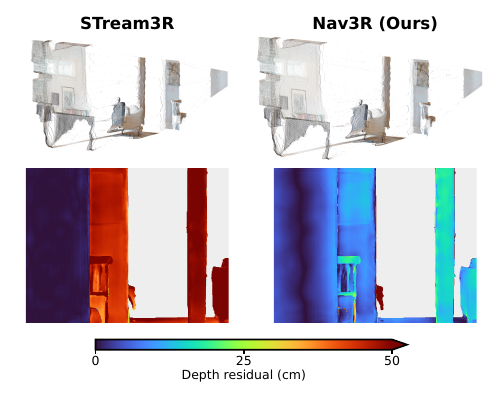}
\caption{\textbf{Qualitative reconstruction comparison after four input frames.} Top: full single-frame predictions, rather than accumulated maps, shown at a shared viewpoint and metric scale using predicted intrinsics, sensor poses, and deployment depth calibration. Bottom: depth residuals on common valid sensor-depth pixels; cooler colors indicate smaller errors. The full predictions include points outside this valid set.}
\label{fig:reconstruction_qualitative}
\vspace{-0.5cm}
\end{figure}

\begin{table}[t]
\caption{\textbf{Geometry, point-feature, and scene ablations on all 1,000 episodes.} SR, SPL, SoftSPL (\%) and DTG (m) at 0.10\,m. Fix./Act.: understanding category overlap (\%) on the fixed and active sets.}
\vspace{-0.3cm}
\label{tab:mechanism_geometry}
\centering
\footnotesize
\renewcommand{\arraystretch}{1.05}
\setlength{\tabcolsep}{2pt}
\begin{tabular*}{\columnwidth}{@{\extracolsep{\fill}}lrrrrrr@{}}
\toprule
Condition & SR $\uparrow$ & SPL $\uparrow$ & SoftSPL $\uparrow$ & DTG $\downarrow$ & Fix. $\uparrow$ & Act. $\uparrow$ \\
\midrule
MTU3D~\cite{zhu2025mtu3d} & 65.4 & 18.01 & 20.56 & 2.239 & 31.2 & 35.6 \\
Raw RGB-D & 69.4 & 19.76 & 21.87 & 2.089 & 35.7 & 43.5 \\
\textbf{Nav3R} & 69.0 & 19.49 & 21.71 & 2.091 & 33.9 & 39.3 \\
\midrule
Coordinate corruption & 65.4 & 18.02 & 20.57 & 2.239 & 31.2 & 35.6 \\
PTv3-zero & 65.4 & 18.02 & 20.57 & 2.239 & 31.2 & 35.6 \\
Feature shuffle & 64.9 & 17.74 & 20.40 & 2.244 & 29.8 & 30.6 \\
\midrule
Wrong scene 1 & 64.3 & 17.72 & 20.35 & 2.311 & 30.6 & 33.1 \\
Wrong scene 2 & 64.2 & 17.69 & 20.46 & 2.291 & 30.4 & 33.7 \\
Wrong scene 3 & 64.6 & 17.77 & 20.42 & 2.310 & 30.4 & 33.1 \\
Three-replica mean & 64.4 & 17.73 & 20.41 & 2.304 & 30.4 & 33.4 \\
\bottomrule
\end{tabular*}
\vspace{-0.3cm}
\end{table}
\subsection{Embodied Navigation}
\noindent\textbf{Protocol and metrics.}
We report SR, SPL, Prox, SoftSPL, and distance-to-goal (DTG). The navigation comparison includes the frozen MTU3D policy, Raw RGB-D accumulation through the same point branch, and Nav3R. All arms share candidate generation and sensor/action protocols. Raw RGB-D samples every second row and column between 0.5 and 5\,m, assigns unit confidence, retains the latest point per 3\,cm voxel, and caps the map at 131,072 points under the same retention policy, point network, verifier, and thresholds. The agent observes $360\!\times\!640$ RGB-D with a $42^\circ$ horizontal field of view at 1.31\,m, moves 0.25\,m or turns $30^\circ$, and performs a target-free $12\times30^\circ$ scan before every decision within a 500-action budget. The navigation mesh is re-baked with the dataset agent settings, and depth uses the simulator's default clip planes. Methods share noise-free poses.

SR requires STOP within 0.10 or 0.25\,m of a target viewpoint. Let $d_0,d_T$ be initial and final geodesic distances, and $L$ the executed path length including the scans. With $\eta=d_0/\max(d_0,L)$, SPL is success times $\eta$ and SoftSPL is $\max(0,1-d_T/d_0)\eta$. Prox omits STOP; DTG is $d_T$.

Navigation intervals use 10,000 scene-bootstrap draws with episode weighting within resampled scenes, and $p$-values a scene-level sign-flip test. The primary contrast, Nav3R minus MTU3D at SR@0.10, was pre-registered; the controls form a separate Holm-corrected family.

\noindent\textbf{Results.}
Nav3R improves SR@0.10 from 65.4\% to 69.0\% ($+3.60$ points, $p=7\times10^{-4}$) and SPL@0.10 by 0.0147 on all 1,000 episodes (Table~\ref{tab:nav1000}). It also improves SR@0.25, Prox, and SoftSPL, reduces DTG by 0.148\,m, and finishes in fewer steps (254 vs.\ 270); the collision rate is unchanged. The pre-registered analysis of the first 752 paired episodes gave $+3.86$ ($[+1.64,+6.46]$) before completion to 1,000. The gain is largest for chairs ($+11.3$, $[+4.9,+18.6]$) and TV monitors ($+4.4$) and absent for plants and sofas, with Raw RGB-D following the same pattern. Re-running 360 episodes per system reproduced all 720 outcomes, and both scene halves give positive estimates ($+1.41$ and $+5.77$). Raw RGB-D accumulation through the same branch reaches 69.4\%: Raw RGB-D minus MTU3D is $+4.00$ ($[+2.20,+5.97]$), and Nav3R minus Raw RGB-D is $-0.40$ ($[-1.60,+0.79]$, $p=0.62$). For navigation, learned reconstruction therefore reaches the privileged reference of noise-free depth accumulation, consistent with the 1.52\,cm median depth error of admitted updates, and the gain comes from the current-scene point branch.

\subsection{Ablation Study}
\label{sec:ablation}

\noindent\textbf{Geometry and semantics.}
Each control changes one element of Nav3R on all 1,000 episodes without retraining. Coordinate corruption independently permutes X, Y, and Z values across points before PTv3, preserving axis marginals. PTv3-zero clears the point features entering the point head, verifier, and retention policy; feature shuffle permutes them across points, preserving their distribution but breaking the feature--location binding. Wrong scene 1--3 replace the encoded map by those of three same-category training episodes at the same decision index in the current camera frame, without updating the protected set; the three-replica mean averages them per episode. Stored maps, admission, occupancy, and candidate generation remain unchanged at that state. For understanding, the controls are applied at the identical fixed-trajectory states and on a new route per control for all 360 active pairs (309--315 with visible goals), with a pre-registered Holm family per protocol; coordinate corruption and PTv3-zero yield no proposal and reproduce MTU3D.

\begin{table}[t]
\caption{Gating and admission ablations on the same 1,000 episodes, with the protocol and units of Table~\ref{tab:mechanism_geometry}.}
\vspace{-0.3cm}
\label{tab:mechanism_gating}
\centering
\footnotesize
\renewcommand{\arraystretch}{1.05}
\setlength{\tabcolsep}{2pt}
\begin{tabular*}{\columnwidth}{@{\extracolsep{\fill}}lrrrrrr@{}}
\toprule
Condition & SR $\uparrow$ & SPL $\uparrow$ & SoftSPL $\uparrow$ & DTG $\downarrow$ & Fix. $\uparrow$ & Act. $\uparrow$ \\
\midrule
\textbf{Nav3R} & 69.0 & 19.49 & 21.71 & 2.091 & 33.9 & 39.3 \\
\midrule
Gate removed & 67.6 & 19.55 & 22.00 & 2.227 & 32.6 & 35.4 \\
Commit every obs. & 68.7 & 19.65 & 21.89 & 2.103 & 35.4 & 41.3 \\
\bottomrule
\end{tabular*}
\vspace{-0.3cm}
\end{table}
\begin{table}[t]
\caption{\textbf{Proposal insertions on the 1,000 episodes.} Counts include inert insertions in the first four decisions. Episodes: episodes with at least one insertion. SR (\%) on those episodes for the condition and for MTU3D on the same episodes; descriptive.}
\vspace{-0.3cm}
\label{tab:insertions}
\centering
\footnotesize
\renewcommand{\arraystretch}{1.05}
\setlength{\tabcolsep}{2pt}
\begin{tabular*}{\columnwidth}{@{\extracolsep{\fill}}lrrrr@{}}
\toprule
Condition & Insertions & Episodes & SR & SR (MTU3D) \\
\midrule
\textbf{Nav3R} & 1,298 & 496 & 92.5 & 85.3 \\
Raw RGB-D & 1,386 & 513 & 93.0 & 85.2 \\
Commit every observation & 1,574 & 599 & 89.8 & 84.3 \\
Gate removed & 3,633 & 915 & 73.2 & 70.8 \\
Feature shuffle & 414 & 237 & 83.1 & 85.2 \\
Wrong scene 1 & 183 & 104 & 38.5 & 49.0 \\
Wrong scene 2 & 176 & 102 & 36.3 & 48.0 \\
Wrong scene 3 & 225 & 127 & 40.2 & 46.5 \\
Coord.\ corruption, PTv3-zero & 0 & 0 & -- & -- \\
\bottomrule
\end{tabular*}
\vspace{-0.3cm}
\end{table}

Every control that breaks the correspondence between point evidence and the current scene removes the gain (Table~\ref{tab:mechanism_geometry}). With Holm correction over the seven controls, Nav3R exceeds the wrong-scene mean by 4.63 SR points ($[+2.89,+6.63]$, $p<10^{-4}$), feature shuffle by 4.10 ($[+2.49,+5.80]$, $p=0.0003$), and coordinate corruption and PTv3-zero by 3.60 ($p=0.004$). Wrong-scene maps are even slightly harmful ($-1.03$ points vs.\ MTU3D, $[-1.73,-0.43]$): the gate admits few of their clusters, but those it admits mislead the policy. Episodes without an insertion match MTU3D's success in every arm, so each SR difference is the net change on episodes with insertions (Table~\ref{tab:insertions}). Understanding responds alike: on fixed and active sets, Nav3R exceeds the wrong-scene mean by 3.47 and 5.81 points, feature shuffle by 4.13 and 8.53, and coordinate corruption and PTv3-zero by 2.66 and 3.69 (all Holm $p<0.005$), and wrong-scene and shuffled features fall below MTU3D. The gain thus requires point features bound to the correct geometry of the current scene, not the insertion mechanism itself.

\noindent\textbf{Geometric reference.}
Raw RGB-D is privileged in simulation: noise-free depth and poses place its points on the rendered surfaces that understanding is scored against. It matches Nav3R in navigation but exceeds it in category overlap by 1.75 ($[+1.17,+2.37]$) and 4.13 ($[+2.66,+5.68]$) points on fixed and active sets, so Nav3R attains 60\% and 47\% of the reference gain over MTU3D. On fixed trajectories the gap is larger in precision ($2.62$ points) than in recall ($1.10$), consistent with centimeter-level reconstruction error that a 10\,cm surface criterion penalizes but a goal insertion rarely does.

\noindent\textbf{Admission and gating.}
Across 94,607 reconstruction proposals from the 1,000 Nav3R runs, 93,565 are admitted. Accepted updates have 1.52\,cm median-of-median and 4.25\,cm median P90 depth error, versus 8.35\,cm and 25.06\,cm for rejected updates. Committing every observation increases accepted writes from 93,565 to 230,338 (2.5$\times$) and leaves navigation unchanged ($+0.30$ points, $[-1.21,+1.69]$; Table~\ref{tab:mechanism_gating}); translation gating removes redundant writes without measurable navigation loss, whereas the additional views raise understanding overlap by 1.44 and 1.96 points (both intervals exclude zero). Removing the gate (no score threshold, persistence, or comparison with the frozen policy) raises insertions from 1,298 in 496 episodes to 3,633 in 915. In navigation, Nav3R's advantage over it is not significant ($+1.40$, $[-1.12,+4.21]$) and its gain over MTU3D loses significance ($+2.20$, $[-1.20,+5.47]$), although SPL still improves; in understanding, it halves the fixed-trajectory gain (Nav3R $+1.35$, $[+0.66,+2.07]$) and eliminates the active one ($+3.95$, $[+1.74,+6.48]$).

\subsection{Qualitative result}
Figure~\ref{fig:realworld} shows a quadruped robot approaching a chair, with egocentric observations, external views, and reconstruction snapshots illustrating how scene geometry is recovered as the viewpoint changes. Figure~\ref{fig:reconstruction_qualitative} compares reconstruction on an HM3D indoor sequence. Nav3R reduces depth residuals over broad visible surfaces in this example, complementing the quantitative reconstruction results.

\section{CONCLUSION}
This paper presented Nav3R, a streaming scene reconstruction framework that turns current-scene geometry into controlled memory for embodied understanding and navigation. Sparse-view priors and causal propagation provide persistent reconstruction, transactional admission and goal-protected retention organize the geometry, and point-semantic proposals with gated insertion connect it to a frozen ObjectNav policy. Pre-registered evaluations on new episodes show that the deployed system designates more of the observed goal surfaces, at matched states and along each policy's own route. On all 1,000 validation episodes, the point memory improves success at the standard 0.10\,m criterion and SPL over frozen MTU3D, and current-scene interventions show that both gains require correct current-scene point evidence. Learned reconstruction matches a privileged noise-free depth reference in navigation and attains about half of its understanding gain, leaving centimeter-level geometric error as the remaining margin. Settings with unreliable depth, instance-level grounding, and quantitative real-robot validation remain directions for extending these results.

\bibliographystyle{IEEEtran}

\bibliography{nav3r_references}
\end{document}